\documentclass[journal]{IEEEtran}
        
\usepackage{cite}
\usepackage{amsmath,amssymb,amsfonts}
\usepackage{algorithmic}
\usepackage{graphicx}
\usepackage{textcomp}
\usepackage{xcolor}
\usepackage{xurl}
\usepackage{booktabs}
\usepackage{float}
\usepackage{adjustbox}
\usepackage{pythonhighlight}
\usepackage{listings}
\usepackage{fancyhdr}
\usepackage{array}
\usepackage[ruled,vlined]{algorithm2e}

\usepackage{array}
\usepackage[caption=false,font=normalsize,labelfont=sf,textfont=sf]{subfig}
\usepackage{stfloats}
\usepackage{verbatim}
\usepackage[most]{tcolorbox}
\usepackage{lipsum} 
\usepackage{orcidlink}
\usepackage[normalem]{ulem}

\makeatletter
\newcommand\footnoteref[1]{\protected@xdef\@thefnmark{\ref{#1}}\@footnotemark}
\makeatother

\begin{document}
\title{
    Large Language Model-Driven Closed-Loop UAV Operation with Semantic Observations
}

\author{Wenhao Wang\orcidlink{0009-0004-7042-341X}, ~\IEEEmembership{Graduate Student Member,~IEEE}, Yanyan Li\orcidlink{0009-0008-2985-572X}, ~\IEEEmembership{Member,~IEEE}, Long Jiao\orcidlink{0000-0002-6790-8932}, ~\IEEEmembership{Member,~IEEE}, Jiawei Yuan\orcidlink{0000-0003-3447-0740}, ~\IEEEmembership{Senior Member,~IEEE}
    \thanks{This work is supported by the US National Science Foundation Award 2318710 and Award 2318711.}
    \thanks{Wenhao Wang,  Long Jiao, and Jiawei Yuan are with the Department of CIS, University of Massachusetts Dartmouth, North Dartmouth, MA 02747 USA {\tt\small (wwang5@umassd.edu; ljiao@umassd.edu; jyuan@umassd.edu)}}%
    \thanks{Yanyan Li is with the Department of CSIS, California State University San Marcos, San Marcos, CA 92096 USA 
            {\tt\small (email: yali@csusm.edu)}}
    \thanks{Copyright (c) 20xx IEEE. Personal use of this material is permitted. However, permission to use this material for any other purposes must be obtained from the IEEE by sending a request to pubs-permissions@ieee.org.}
}

\maketitle \thispagestyle{fancy}

\begin{abstract}
    Recent advances in Large Language Models (LLMs) have revolutionized mobile robots, including unmanned aerial vehicles (UAVs), enabling their intelligent operation within Internet of Things (IoT) ecosystems. However, LLMs still face challenges from logical reasoning and complex decision-making, leading to concerns about the reliability of LLM-driven UAV operations in IoT applications. In this paper, we propose a closed-loop LLM-driven UAV operation code generation framework that enables reliable UAV operations powered by effective feedback and refinement using two LLM modules, i.e., a Code Generator and an Evaluator. Our framework transforms numerical state observations from UAV operations into semantic trajectory descriptions to enhance the evaluator LLM's understanding of UAV dynamics for precise feedback generation. Our framework also enables a simulation-based refinement process, and hence eliminates the risks to physical UAVs caused by incorrect code execution during the refinement. Extensive experiments on UAV control tasks with different complexities are conducted. The experimental results show that our framework can achieve reliable UAV operations using LLMs, which significantly outperforms baseline methods in terms of success rate and completeness with the increase of task complexity.
\end{abstract}


\section{Introduction}\label{sec:intro}
\begin{figure}[t]
    \centering
    \includegraphics[width=1\linewidth]{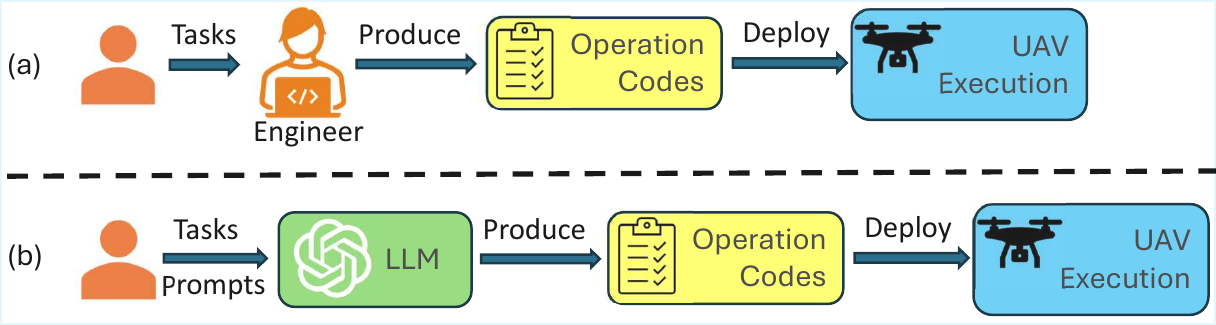}
    \caption{Pipelines of robotic operation code generation: (a) without LLMs and (b) with LLMs.}
    \label{fig:piplines}
\end{figure}

\IEEEPARstart{U}{avs}, or drones, are increasingly being integrated into IoT ecosystems for various applications, such as surveillance and monitoring, smart cities, and infrastructure inspection \cite{uav_iot_applications}. Their high mobility and rich sensing capabilities make them an ideal candidate for data collection, communication, and service delivery tasks in these applications. Given a UAV task, the traditional pipeline requires a specialized engineer to write the operation code based on the task description and then deploy the code on the device for execution, as shown in Fig.~\ref{fig:piplines}(a). Such a process requires substantial domain knowledge and can also be time-consuming. With LLMs' exceptional semantic understanding and context generation capabilities, recent research has demonstrated the potential of leveraging LLM to facilitate the operation code generation process for a given robotic task \cite{ChatGPTRobotics, TypeFly, CLEAR, GSCE, incoro, InteractivePlanning, CoPAL, CAPE, SelfRefine, trust}. As illustrated in Fig.~\ref{fig:piplines}(b), these LLM-driven approaches enable the generation of UAV operation code with natural language tasks, which significantly simplifies the code generation and also brings the potential of automated UAV task deployment.

Despite the benefits brought by LLM-driven UAV operation code generation, it also raises concerns about the reliability of LLM-generated code due to the challenges faced by LLM in terms of logical reasoning and complex decision-making \cite{UAVsmeetllmsoverviews}. Different from text generation applications, where semantic-level accuracy is sufficient, the generation of UAV operation code requires the precise execution of sequential actions. Erroneous or invalid operation code can lead to unsafe UAV behavior and pose safety and security risks to the public, such as UAV crashes. 

To enhance the reliability of LLM in translating natural language tasks into operation code, techniques such as prompt engineering \cite{PromptEngineering} and few-shot learning \cite{few-shot} have been widely adopted when designing LLM-driven operation code generation research for robots, including UAVs, and have demonstrated promising results. However, a set of existing approaches \cite{ChatGPTRobotics, TypeFly, CLEAR, GSCE} adopt an open-loop design that lacks mechanisms to verify the correctness of the operation code generated by LLMs. Such open-loop designs can leave questions such as cumulative error and temporal action consistency unsolved \cite{OpenLoopQuestions}, and hence affect the reliability of their generated code, especially when handling complex tasks. Recently, closed-loop designs have been proposed in LLM-driven robotic operation code generation \cite{incoro, InteractivePlanning, CoPAL, CAPE, SelfRefine, trust}, which enable evaluation of generated code and provide feedback for refinement, thereby steering robot execution towards its task objectives. Nevertheless, these approaches still face limitations from different aspects, including:
\begin{itemize}
    \item Using numerical state observations for evaluation: approaches proposed in ref \cite{incoro, InteractivePlanning} use numerical state observations in scenarios for evaluation and feedback generation. Due to the relatively limited numerical reasoning and processing capabilities of LLMs \cite{NumberUnderstanding}, an LLM-driven evaluation can overlook the latent semantics of robot actions and fail to achieve a comprehensive assessment of its movements and dynamics, leading to unreliable code refinement feedback.
    \item Physical execution for evaluation: closed-loop approaches, such as ref \cite{CoPAL, CAPE, InteractivePlanning}, employ physical execution of LLM-generated code to generate state observations for evaluation and code refinement. Although these approaches can be acceptable for robots that are not likely to cause physical damage during the refinement, their adoption in the context of UAV operation becomes risky because the physical execution of UAVs with code that potentially contains errors for refinement can easily lead to UAV crashes and safety concerns \cite{uav-crash}. 
    \item Self-refinement using a single LLM: existing closed-loop approaches \cite{SelfRefine, trust} that a single LLM can be used for both code generation, evaluation, and refinement. However, self-evaluation may amplify its errors and lead to degradation in task success rate \cite{LLM-bias}, which is also validated in the experimental evaluation of this work.
\end{itemize}

To overcome these limitations, this paper proposes a novel closed-loop framework to further enhance the reliability of LLM-driven UAV operation code generation. Specifically, our framework adopts a duo-LLM design, in which one LLM is configured as the \emph{Code Generator} and the other one is configured as the \emph{Evaluator}. To compensate for LLM's limitation in numerical reasoning, we propose a state transformation algorithm that converts numerical state observations into semantic observations by embedding UAV dynamics within the context of semantic trajectory observations during closed-loop evaluation. In addition, our framework enables simulation-based evaluation and refinement of LLM-generated UAV operation code before physical deployment. This feature enables efficient refinement for UAV operation code generation, and more importantly, eliminates the safety risks raised by potential incorrect code execution during refinement with physical UAV execution. Align with existing research \cite{incoro, InteractivePlanning, CoPAL, CAPE, SelfRefine, trust}, this paper also focuses on using LLMs to generate operation code for deployment and task execution with the complete task description provided, but not real-time interactive UAV controls.

\begin{figure}[t]
    \centering
    \includegraphics[width=1\linewidth]{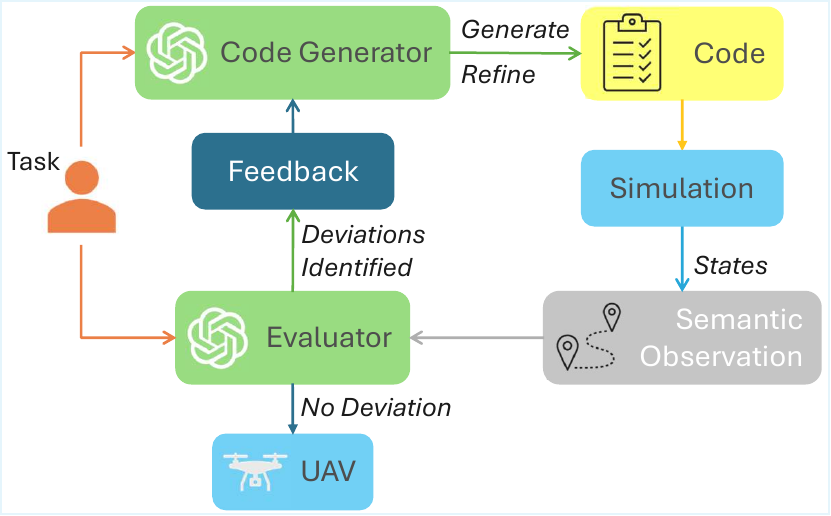}
    \caption{Illustration of LLM-driven closed-loop UAV Operation and Refinement.}
    \label{fig:overall}
\end{figure}

The high-level design of our framework is illustrated in Fig.~\ref{fig:overall}. In our framework, the LLM configured as the \emph{Code Generator} synthesizes or refines UAV operation code based on task description or evaluation feedback, and the LLM configured as the \emph{Evaluator} assesses code performance and provides feedback to guide code refinement. The process unfolds as follows: 1) The \emph{Code Generator} first produces an initial code based on the task description; 2) The code is then executed in simulation to produce raw numerical state observation, which are subsequently transformed into semantic observations; 3) The \emph{Evaluator} evaluates the semantic observations against the original task specification and returns evaluation feedback; 4) If the evaluation confirms that the observations have met the task objectives, the code will deploy on the UAV for task execution. Otherwise, the evaluation feedback is returned to the \emph{Code Generator} for iterative refinement. 

To evaluate the effectiveness of our proposed framework, we conducted extensive experiments on UAV operation tasks with varying levels of complexity. The experimental results show that our framework achieves a $100\%$ success rate for basic tasks. For advanced tasks with a higher complexity, our framework achieves a $98.5\%$ completeness and an $85\%$ success rate, i.e., $98.5\%$ of required actions in all evaluated tasks are completed correctly and $85\%$ of tasks are entirely completed without any error. Compared with baseline approaches \cite{GSCE,SelfRefine,InteractivePlanning}, our framework greatly improves both success rate and completeness when the task complexity increases.

The remainder of this paper is structured as follows: Section \ref{sec:related work} reviews and discusses existing research. Section \ref{sec:problem formulation} presents our problem formulation, which is followed by the detailed design of our proposed framework in Section \ref{sec:method}. Section \ref{sec:experiment} provides experimental evaluation and analysis of our framework. We discuss the applicability of our framework and future work in Section \ref{sec:discussion}, and conclude this paper in Section \ref{sec:conclusion}.

\section{Related Work}\label{sec:related work}
\subsection{LLM for Robotics}
Continual advancements in LLMs \cite{deepseek, o3mini} have demonstrated their exceptional capabilities in semantic understanding and context generation, making them promising when integrating with various robotics tasks. Research works from Microsoft \cite{ChatGPTRobotics} and Google \cite{CodeasPolicies} demonstrated how publicly accessible LLMs can be leveraged to generate operation code for various robotic tasks, including UAVs. Since then, a set of research works \cite{TypeFly, CLEAR, GSCE, incoro, InteractivePlanning, CoPAL, CAPE, SelfRefine, trust} have been proposed to enhance the quality and reliability of LLM-driven robotic tasks, especially with the integration of prompt engineering techniques \cite{PromptEngineering} that help steer LLMs' behavior and align their outputs with the tasks' objectives.

In ref \cite{TypeFly}, an end-to-end LLM framework is proposed that enables semantic planning with applications in quadcopter drones. This paper also introduced a specific programming language (MiniSpec) to improve the code generation efficiency of LLMs. In ref \cite{CLEAR}, a prompt-engineered robot control solution is proposed that synthesizes input from both LLMs and humans. This work aims to have the robotic system behavior programmed by emphasizing prompting LLMs and avoiding manipulating code. Recently, a prompt framework with enhanced reasoning, named GSCE \cite{GSCE}, was proposed to better support LLM-driven UAV control. GSCE integrates domain-specific guidelines, constraints, and examples into its system prompts when configuring the LLM, and hence improves its reliability of constraint-compliant code generation. While these methods all improve the LLM-driven code generation for robotic tasks from different aspects, they all adopt an open-loop design framework without error correction and adjustment for the generated code. Such a design can lead to potential cumulative errors and temporal action consistency issues \cite{OpenLoopQuestions} in the code generation.

To address the concerns from the open-loop design, different closed-loop refinement strategies are increasingly adopted by recent research towards LLM-driven code generation for robotic tasks. In ref \cite{incoro, InteractivePlanning}, numerical observation of the robot's execution status, such as actuator reading, sensor reading, and system information (e.g., location), is utilized to construct the feedback for LLMs. These numerical observations-based feedback can positively contribute to the quality of LLM-generated operation code, however, their contribution is restricted when trying to assess the latent semantics of robot actions for a comprehensive assessment. This is because LLMs have relatively limited numerical reasoning and processing capabilities \cite{NumberUnderstanding}. In addition, to obtain accurate observations and leverage them for the refinement of LLM-generated code, executing the code with physical experiments is employed for feasible robotic tasks. For example, CoPAL~\cite{CoPAL} employs physical error feedback and run-time error feedback to refine the code when errors occur during physical execution. In ref \cite{CAPE}, precondition errors are integrated into a prompt that aims to repair plans when the robot fails skill execution. While evaluating code with physical experiments can provide accurate observations of errors and misalignment between LLM-generated code and task objectives that enable the construction of an effective refinement plan, it does not fit when moving to the context of many UAV tasks. Physical execution of code that may contain errors, especially before any refinement, has a high chance of causing the UAV to crash \cite{uav-crash}. This not only increases the hardware cost for the experiment but also can lead to safety concerns. Therefore, our proposed framework enables the execution of LLM-generated UAV operation code in the simulation environment to resolve the safety concerns in the refinement process. Moreover, the state transformation design in our framework enables semantic observations of the execution rather than numerical-only observation, and hence empowers LLMs in the code evaluation and feedback generation for refinement.

In recent research, other strategies to further improve the performance of closed-loop design have also been explored. For example, ref \cite{preemptive} utilizes human feedback to revise the code when misalignment in the generated plan is detected. Although human feedback provides high-quality input for refinement, it limits the automation of the system and the performance depends on the expertise of the user. In addition, existing research \cite{incoro, InteractivePlanning, CoPAL, CAPE} separates the configuration of the code generator and evaluator on different LLMs. This design aims to minimize the amplification of errors and bias \cite{LLM-bias} that can be caused by using a single LLM for both code generation and evaluation in previous research \cite{SelfRefine, trust}. In our framework, we also use two separate LLMs and have customized configurations to improve their capabilities as the code generator and evaluator, respectively.

\subsection{Prompt Engineering}
Recent studies have increasingly employed prompt engineering~\cite{PromptEngineering} to improve the reliability of LLM-driven robotic systems. For instance, GSCE proposes a prompt framework as a system prompt to improve the reliability of LLM-driven UAV control code generation~\cite{GSCE}. Prompt engineering typically leverages two core techniques: in-context learning~\cite{URIAL} and Chain-of-Thought (CoT) reasoning~\cite{CoT}. In-context learning enables the LLM to identify patterns from a limited set of examples and apply them to new situations. This facilitates the generation of control code that adheres to robot policy and how to ground task descriptions from a few examples~\cite{CodeasPolicies, PromptBook}. On the other hand, CoT prompts the LLM to articulate intermediate inference steps, which is valuable for robotic tasks requiring sequential, stepwise decision-making. By decomposing a task into steps, CoT encourages the production of code that aligns with each stage of the intended action plan. Prior studies have embedded CoT within the examples to guide LLMs through reasoning step-by-step via in-context learning~\cite{ISR-LLM, PromptBook}. In this study, we integrate both CoT and in-context learning strategies to enhance the quality of code generation and evaluation.

\section{Problem Formulation} \label{sec:problem formulation}
Given a UAV operation task description, the framework proposed in this paper aims to use LLMs to reason through the task description and generate accurate operation code after observation, evaluation, feedback, and refinement. Specifically, we formulate our problem by extending the LLM-driven UAV operation problem in \cite{GSCE} as \(P = \langle s_0, S, A, T, G \rangle\), where \(s_0\) is the initial state, \(S = (s_0, s_1, s_2, \dots, s_n) \) is the finite set of discrete states encoding the UAV's spatial position and orientation, \(A\) is the finite action (UAV skills) set, \(T\) is the deterministic transition, and \(G \subseteq S\) is the goal state. For a current state \(S\), an action \( a \in A \) is selected from the skill set \( A \), leading to a deterministic state transition: \(T: S \times A \rightarrow S\) that determines the next state.  A candidate solution to problem \(P\) is a sequence of actions \(l = (a_1, a_2, \dots, a_n)\) that control the UAV from the initial state \( s_0 \) through intermediate states until reaching a final state \(s_n\). The state transitions of \(l\) yield state observations \(O\), and evaluation \(E = \langle  O, P \rangle\) assesses whether \(O\) satisfies the problem objectives and outputs \(F\) as feedback. If \(F\) is ``YES'', the sequence of actions \(l\) is deemed a valid solution that flies the UAV from \(s_0\) to the goal state \((s_n \equiv G\)); otherwise, \(l\) is iteratively refined based on the feedback \(F\).

\begin{figure*}[!htbp]
    \centering{\includegraphics[scale=0.75]{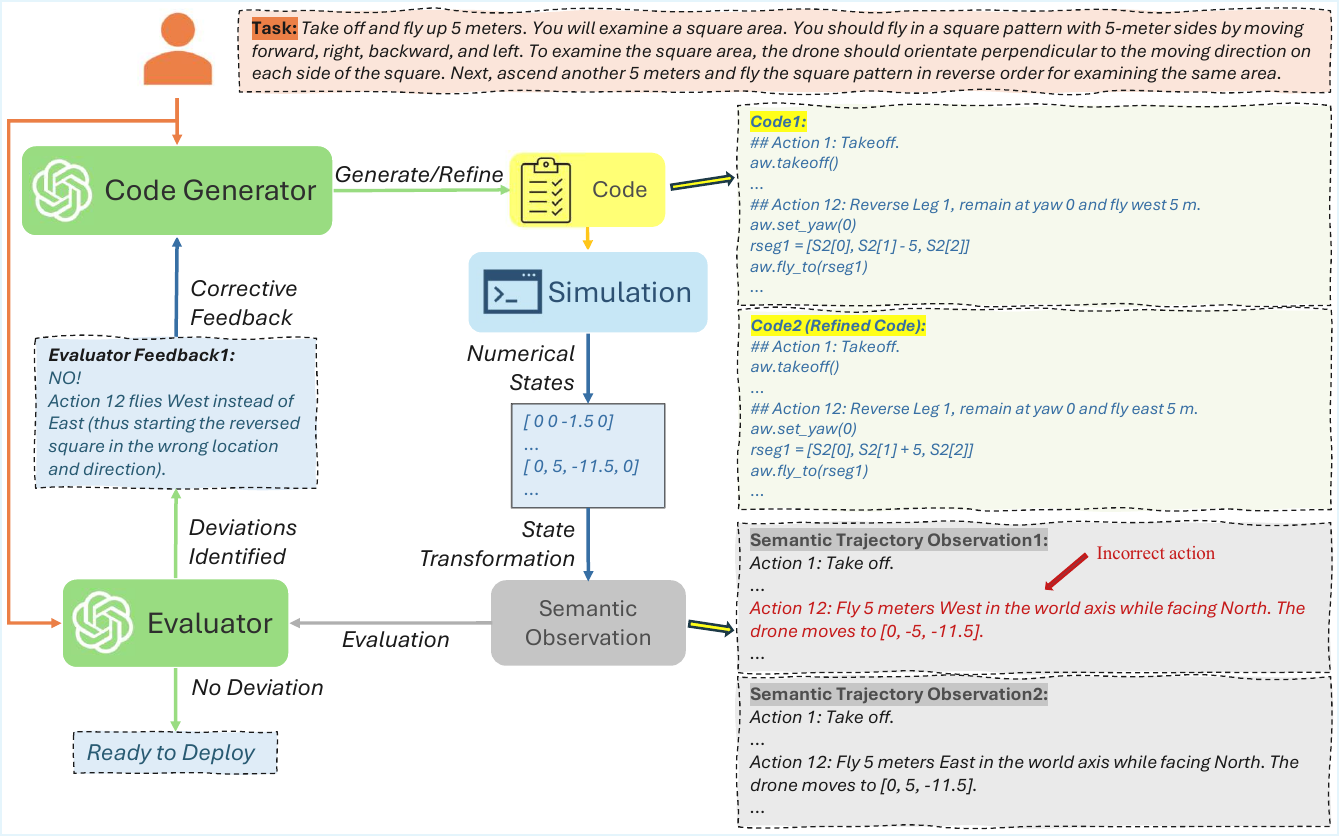}}
    \caption{Overall design of LLM-driven closed-loop UAV operation code generation with semantic observations. An example task of examining a square area is presented. In the first loop, the evaluator identifies deviations in the semantic trajectory observation (action 12), providing error feedback into the code generator to refine the code. In the second loop, the evaluator confirms refined code matches the task, and the code is deployed on the UAV for mission execution.
    }
    \label{fig:feedback}
\end{figure*}

\section{Methodology} \label{sec:method}

\subsection{Overview} \label{sec:overview}
The overall design of our proposed reliable LLM-driven closed-loop control framework for UAV operations is presented in Fig. \ref{fig:feedback}, in which an example task of examining a square area is also added. In addition, we also summarize the closed-loop feedback and refinement process of our LLM-driven code generation in Algorithm~\ref{alg:overall}. Specifically, given a UAV operation task description (\(P\)) by the user, an LLM configured as the code generator with enhanced reasoning capabilities towards UAV operation first produces the initial version of the operation code \(l\). A simulation-based execution of the code \(l\) is then performed to obtain the numerical state observations. To close the gap between numerical state observations and the LLM-driven evaluation, a state transformation is designed that transforms numerical state observations into a semantic trajectory observation \(O\). After that, the evaluator analyzes the semantic observation together with the task description and output evaluation feedback \(F\) that depict the deviations from the \(P\) in \(O\) (if any). If the trajectory \(O\) satisfies all task objectives in evaluation, the operation code is ready for deployment; otherwise, the code generator re-generates the refined \(l\) according to the evaluator's feedback \(F\). This ``generation-simulation-evaluation" process iterates until the evaluation confirms task objectives are achieved or the maximum number of iterations is reached. We now provide the detailed design of our code generation, trajectory observation, evaluation,  as well as feedback and refinement in the following sections.

\begin{algorithm}[!htbp]
    \SetAlgoLined
    \caption{Closed-loop Feedback and Refinement}\label{alg:overall}
    \KwIn{$P$: Task description}
    \KwOut{$l$: Code script}

    $l \gets$ Code Generator$(P)$

    $itr \gets 1$

    \While{$itr \leq$ maximum round}
    {   

        $O \gets$ State Transformation$(l)$

        $F \gets$ Evaluator$(O, P)$

        \eIf{``YES" in $F$}
        {
            \textbf{break}
        }
        {
            $l \gets$ Code Generator$(F)$
        }

        $itr \gets$ $itr + 1$

    }

    \Return $l$

\end{algorithm}

\subsection{Code Generation} \label{sec:code generation}
The LLM-driven code generation in our framework is achieved by configuring a general-purpose LLM agent (code generator) with UAV operation-related system prompts. We adopt prompt engineering strategies proposed by the GSCE framework \cite{GSCE}, which have been demonstrated to be effective for enhancing the reasoning capabilities of LLMs in basic UAV operation code generation. Specifically, the system prompts comprise guidelines, skill APIs, constraints, and examples. The guidelines shape the LLM's role for code generation, the skill APIs empower LLM to employ appropriate functions, the constraints regulate LLM with robotic policies, and the examples are pairs of query-response that serve as illustrations. These strategies serve as the basic building blocks to support the LLM-driven UAV operation code generation. On top of that, our code generation also utilizes constructive feedback from our evaluation design in section~\ref{sec:evaluation}. The feedback helps the code generator obtain a more comprehensive understanding of the dynamics and interconnections among UAV operations implied by the task description, which enhances the LLM's reasoning for code generation and code refinement.

\subsection{Trajectory Observation}\label{sec:trajectory observation}
Trajectory observation constitutes the foundation of our closed-loop feedback. It embeds the observations of the UAV's historical trajectory dynamics, thereby enabling the evaluator to identify its deviations from the task objectives. Rather than getting trajectory observations from real robots~\cite{InteractivePlanning, CoPAL} that can cause a UAV crash if the LLM-generated code contains errors, our framework adopts simulation-based execution to generate trajectory observations. When the LLM-generated code is executed in simulation, it produces the UAV's state observations for each action. These observations are numerical representations of the UAV's states, including the UAV's position and orientation. However, numerical observations offer limited semantic insight for an LLM-based evaluator because LLM lacks prior knowledge about interpreting the semantic meaning of the numerical state vector \cite{NumberUnderstanding}. 

\begin{figure}[!htbp]
    \centering
    \includegraphics[width=.9\linewidth]{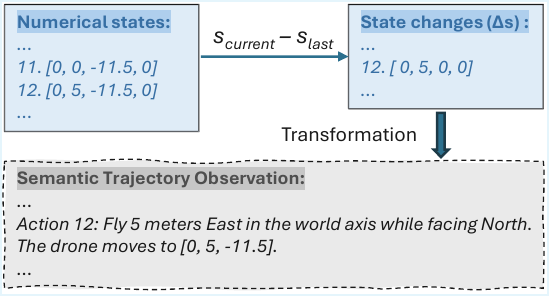}
    \caption{An example of transforming numerical state observations into semantic trajectory observation. UAV state vector of \([x, y, z, \theta]\) denotes (North, East, Down, Yaw).}
    \label{fig:state transformation}
\end{figure}

To bridge this gap, we transform each numerical state observation into a step-wise semantic description and aggregate these into a semantic trajectory observation. As an example shown in Fig.~\ref{fig:state transformation}, the semantic description captures the semantic meaning of the state changes and UAV position, such as ``Move 5 meters East while facing North, the UAV moves to [0, 5, -11.5]". This transformation provides the evaluator with a clear understanding of the semantic significance of each action and overall trajectory dynamics, thereby improving the evaluator's ability to identify deviations and generate precise feedback for guiding the code generator on code refinement.

Algorithm~\ref{alg:trajectory observation} presents the transformation from numerical state observations to the semantic trajectory observation \(O\). For each action identified in the \(l\), its execution in the simulation will be observed to obtain the corresponding numerical UAV state observations \([x, y ,z, \theta]\) and then transformed to semantic trajectory descriptions as the example illustrated in Fig.~\ref{fig:state transformation}. Specifically, for each action \(a\), the simulator first executes \(a\), if a runtime error occurs, the algorithm collect and log errors alongside the corresponding action identifier in \(O\); otherwise, the algorithm computes the state transition \(T\) between the current state \(s_{\mathrm{current}}\) and the last state \(s_{\mathrm{last}}\), denoted as \(T = s_{\mathrm{current}} - s_{\mathrm{last}}\). For each element in the state transition \(T\), the algorithm appends the corresponding semantic description of orientation and changes in yaw, x-axis, y-axis, or z-axis to \(O\). The algorithm also appends the UAV's NED position vector to \(O\) for embedding the positional information. Upon completion of all actions of \(l\), the algorithm returns the complete semantic trajectory observation \(O\).

\begin{algorithm}[!htbp]
    \SetAlgoLined
    \caption{State Transformation}\label{alg:trajectory observation}
    \KwIn{$l$: Code script}
    \KwOut{$O$: Semantic trajectory observation}

    $O \gets \emptyset$

    \ForEach{ $action$ in $l$ }
    {
        $s_{\mathrm{last}} \gets$ Numerical state observation $[x, y, z, \theta]$

        Execute $action$

        \If{error executing $action$}
        {
            $O \gets O $ + ``Error in executing $action$ with error message $error$"
        }

        $s_{\mathrm{current}} \gets$ Numerical state observation $[x', y', z', \theta']$

        $T \gets s_{\mathrm{current}} - s_{\mathrm{last}}$

        \ForEach{$element$  in $T$}
        {
            \eIf{$element$ in orientation}
            {
                $O \gets O$ + Rotate $element$ degrees in Yaw.
            }
            {\If{$element$ in NED axis}
                {
                    $O \gets O$ + ``Move $element$ meters in North, East, or Down while facing $s_{\mathrm{current}}[\theta]$."
                }
            }

            $O \gets O$ + ``The UAV moves to $s_{\mathrm{current}}[x, y, z]$."
        }
    }
    \Return $O$
\end{algorithm}

\subsection{Evaluation} \label{sec:evaluation}
In our framework, we design the evaluation by leveraging the LLM's semantic reasoning capabilities to analyze the semantic trajectory observation. The evaluator aims to identify deviations (if any) between the semantic trajectory observation and the task description, and then construct feedback based on the deviations. Instead of using the same LLM used for code generation for self-feedback evaluation, our design employs a separate LLM to prevent the same LLM from amplifying its bias~\cite{LLM-bias}.

\begin{figure}[!htbp]
    \centering
    \includegraphics[width=.9\linewidth]{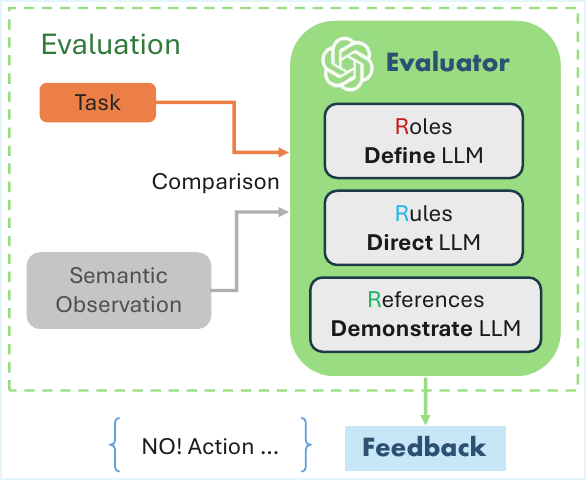}
    \caption{The evaluation process begins by configuring the LLM agent (evaluator) via a system prompt framework that includes roles, rules, and references. Next, the semantic observations and the task description are provided to the evaluator for comparison.}
    \label{fig:evaluator}
\end{figure}

To improve the accuracy of LLM's evaluation, we adopt the design principles of the code generator by configuring the evaluator as an LLM agent with a structured system prompt. However, unlike the code generator, the evaluator does not produce code but instead generates evaluation feedback. To address this issue, we construct the evaluator's system prompt with \emph{roles}, \emph{rules}, and \emph{references}, as depicted in Fig.~\ref{fig:evaluator}. Specifically, the \emph{roles} define the LLM agent as an evaluator that compares the semantic trajectory observation against the task description, while the \emph{rules} serve as explicit evaluation criteria to direct the LLM when reasoning about trajectories. Additionally, the \emph{references} are example trajectories that serve as benchmarks to support the evaluation of semantic trajectory observations. Furthermore, the system prompt requires the evaluation outcome to be informative and constructive by providing explanations about the depicted deviations. As demonstrated in Fig.~\ref{fig:feedback}, given a semantic trajectory observation and a task description, the structured system prompt enables the LLM-evaluator to identify deviations from the task description and explain the errors in the trajectory observation (e.g., ``Action 12 flies West instead of East''). Consequently, the code generator is aware of the errors and could correctly refine the code in the subsequent refinement process.

\subsection{Feedback and Refinement} \label{sec:feedback and refinement}
The feedback and refinement steer the LLM-generated UAV operation code towards fully satisfying the task objectives. They leverage the two LLM agents designed in section~\ref{sec:code generation} and section~\ref{sec:evaluation} and form a closed-loop as illustrated in Fig.~\ref{fig:feedback}. Specifically, the code generator interprets the evaluation feedback \(F\) and refines the generated code accordingly. Then, the refined code is executed in the simulation again to produce a new semantic trajectory observation \(O'\) for re-evaluation. The feedback and refinement loop continues iteratively until the evaluator outputs ``YES", indicating that \(E\) confirms that the refined action sequence \(l\) successfully solves the task problem \(P\) and the UAV operation code is ready for deployment. In addition, if the maximum number of iterations is reached without a successful evaluation, human supervision should be involved to address potential errors in the code.

\section{Experiment and Evaluation} \label{sec:experiment}
\subsection{Experimental Setup} \label{sec:experimental setup}
In our experiment, the implementation of our proposed framework and comparison methods \cite{GSCE, SelfRefine, InteractivePlanning} uses the OpenAI ``o3-mini'' (o3-mini-2025-01-31) model \cite{o3mini} as the foundational LLM. Based on the ``o3-mini'', two separate LLMs are configured as the code generator and evaluator in our framework. The widely used UAV simulator, ``AirSim'' \cite{AirSim}, is tailored to support the simulation component of our framework for code execution and state observations. The UAV model selected for experiment and evaluation is the default quadcopter in AirSim. The detailed implementation of our framework and experiment is available at our project website \footnote{\label{note1}Project Website: \url{https://github.com/ai-uavsec/CLGSCE}}.

\underline{UAV Task Dataset}: In our experiment, two categories of UAV tasks are selected to evaluate the performance of our framework under different operational complexities, i.e., \textbf{Basic} tasks and \textbf{Advanced} tasks. Each \textbf{Basic} task involves one or two intuitive actions to evaluate LLM's few-stage planning and basic spatial reasoning (e.g., \textit{``Turn 90° clockwise, then fly 4 meters left in the drone's body frame''}). We use the \textbf{Basic} task dataset established by existing research \cite{GSCE}, which contains 44 tasks in total. Differently, the \textbf{Advanced} tasks in our experiment require complex reasoning about the UAV's position and orientation dynamics and typically require 6-19 actions to reach the goal state. We established the \textbf{Advanced} task dataset by emulating real-world UAV operation tasks, which consists of 20 tasks with different complexities. As summarized in Table~\ref{tab:advanced tasks}, there are four levels of complexity: ``Low'', ``Medium'', ``High'', and ``Very High''. The ``Low'' tasks involve flying simple geometric patterns. The ``Medium'' tasks introduce orientation requirements while flying the patterns (e.g., always orienting to the flying direction). The ``High'' tasks require flying complex patterns with orientation requirements (e.g., fly a figure 8 while orienting in the flying direction). The ``Very High'' increases the complexity by adding requirements to the patterns (e.g., inspect an area twice with the second inspection flying the path reversely). The complete list of both Basic and Advanced tasks is available at our project website~\footnoteref{note1}. All tasks were manually validated in our experiment to avoid potential simulator-induced errors that could affect the evaluation process.

\newcolumntype{L}[1]{>{\raggedright\arraybackslash}p{#1}}
\newcolumntype{C}[1]{>{\centering\arraybackslash}m{#1}}
\newcolumntype{R}[1]{>{\raggedleft\arraybackslash}p{#1}}

\begin{table*}[!htbp]
    \centering
    \caption{Examples of Tasks in Advanced Task Dataset}
    \begin{tabular}{c c c p{11.5cm}}
        \toprule
        \textbf{Complexity} & \textbf{Num. of Tasks} & \textbf{Avg. Steps} & \multicolumn{1}{c}{\textbf{Example Tasks}}\\
        \midrule
         Low & 3 & 9  & ``You are going to scan square areas. First, take off and climb up 5 m. Fly a 5 m square (forward/right/back/left). Then shift 5 m south, and fly the same square pattern again."\\
        \midrule
        Medium & 9 & 11 & ``You are going to view the borders of square areas. Please take off and fly up 5 meters. You should fly in a square pattern with 5-meter sides by moving forward, right, backward, and left in world axis. Make sure the drone is oriented to the flying direction." \\
        \midrule
        High & 4 & 17.5 & ``You are going to fly a figure of 8. Please take off and fly up 5 meters. Fly a figure of 8 on a flat, horizontal plane with each side of 5 meters. The left square is on your left-rear side, and the right square is on your right-front side. You should begin with the left square by flying left. The right square should start from moving forward. Make sure the drone is oriented in the flying direction."\\
        \midrule
        Very High & 4 & 17.5 & ``You are going to examine a cube. Take off and fly up 5 meters. You will examine a square area. You should fly in a square pattern with 5-meter sides by moving forward, right, backward, and left. To examine the square area, the drone should orientate perpendicular to the moving direction on each side of the square. Next, ascend another 5 meters and fly the square pattern in reverse order to examine the same area."\\
        \bottomrule
    \end{tabular}
    \label{tab:advanced tasks}
\end{table*}

\underline{Baseline Methods for Comparison}: Three recently proposed methods \cite{GSCE,SelfRefine,InteractivePlanning} are selected as the baselines for comparison with our proposed approach, including 

\begin{itemize}
    \item \textbf{GSCE} \cite{GSCE}: An open-loop approach implemented by configuring a general LLM as UAV operation code generation agent (GSCE) with the system prompt that contains guidelines, skill APIs, constraints, and examples. The GSCE is first initialized with the system prompt; upon receiving a task prompt, it synthesizes control code based on the task instructions without any refinement.

    \item \textbf{LLM Self-feedback (Self-Refine)} \cite{SelfRefine}: A single LLM responsible for both code generation and refinement, which extends the GSCE method by self-evaluation and self-refinement of the code. Specifically, after the initial code is generated, it is executed in simulation to produce semantic trajectory observations, then the GSCE is prompted to evaluate these observations against the task description and iteratively refine the code based on its self-assessment.

    \item \textbf{Feedback with Numerical State Observations (Numerical) }\cite{InteractivePlanning}: Similar to our method, this method introduces a secondary LLM that evaluates the UAV states. In implementation, the code generator (GSCE) generates the initial code, which is then executed in simulation to produce UAV state observations in numerical representations of UAV state changes (format is the same as ground truth in the section~\ref{sec:metrics}). Then, a separate LLM is configured as the evaluator with \textit{roles}, \textit{rules}, and \textit{references (numerical trajectory exemplars)}. The evaluator analyzes these numerical observations against the task description and provides corrective guidance to the code generator for subsequent refinement.
\end{itemize}

\subsection{Evaluation Metrics} \label{sec:metrics}
In our evaluation, two key metrics \underline{Completeness} and  \underline{Success Rate} (SR) are used for performance measurement. Given a UAV task, the completeness measures the ratio of actions required by the task that are completed successfully, and the SR measures whether the entire task is completed successfully. The detailed definition of these two metrics, as well as the ground truth used in our evaluation, is provided below.

\underline{Completeness}: As formalized in Eq.~\ref{eq:completeness_i}, completeness for task \(i\) computed as the ratio between the count of actions correctly executed (\(a^{\mathrm{correct}}_i\)) and the total number of actions in the task's ground truth action sequence (\(l^{\mathrm{gt}}_i\)). This metric offers insights into performance across the intermediate control process, where a higher completeness indicates a greater proportion of the task that has been successfully completed.

\begin{equation}
    Task\_Completeness_i = \frac{|a^{\mathrm{correct}}_i|}{|l^{\mathrm{gt}}_i|}
    \label{eq:completeness_i}
\end{equation}

Given \(n\) tasks, the completeness is calculated as:

\begin{equation}
    Completeness = \frac{1}{n}\sum_{i=1}^{n}Task\_Completeness_i
    \label{eq:completeness}
\end{equation}

\underline{SR}: As formalized in Eq.~\ref{eq:success}, a task is counted as a success only when the UAV reaches the final state while following the correct sequence of actions that lead to the ground truth trajectory. Then we compute the SR across \(n\) tasks as Eq.~\ref{eq:SR}. This metric is grounded in real-world operational requirements, wherein the UAV must perform the task precisely as instructed.

\begin{equation}
    Success_i =
    \begin{cases}
        1, & Task\_Completeness_i = 1,    \\
        0, & Task\_Completeness_i < 1.
    \end{cases}
    \label{eq:success}
\end{equation}
\begin{equation}
    SR = \frac{1}{n}\sum_{i=1}^{n}Success_i
    \label{eq:SR}
\end{equation}

\underline{Ground Truth}: The ground truth is represented by a list of state transitions. It is a vector of four elements: \([x, y, z, \theta]\), where \(x\), \(y\), and \(z\) denote the UAV's position changes in the North, East, and Down axes, and \(\theta\) represents yaw rotation. We assume the UAV's state is fully observable at all times, thereby ensuring that each transition can be precisely monitored and compared against ground truth.

\underline{Tolerance}: These metrics evaluate whether the generated code can correctly operate the UAV to perform the actions required by the task description. As AirSim simulation injects white noise in sensor measurements to emulate real-world sensing behavior \cite{AirSim}, there are offsets in the UAV state estimates produced by AirSim. To filter such noises that could affect trajectory matching with ground truth, we set a tolerance threshold of 1 meter for each positional parameter \([x, y, z]\) and 4 degrees for the Yaw \(\theta\) in our implementation.

\subsection{Result and Analysis} \label{sec:result and analysis}
\subsubsection{\textbf{Overall Result}} \label{sec:overall result}
The average SR and completeness across Basic and Advanced tasks are summarized in Table~\ref{tab:overall results}. The results reflect the average value of three repetitions of experiments to mitigate the impact of potential randomness of LLM's output~\cite{LLMrandomness}.

\begin{table}[!htbp]
    \centering
    \caption{SR and Completeness of Proposed Methods}
    \begin{adjustbox}{width=0.48\textwidth}
        \begin{tabular}{lcccc}
            \toprule
                                                 & \multicolumn{2}{c}{Basic Task} & \multicolumn{2}{c}{Advanced Task}                                         \\
            \midrule
                                                 & SR                            & Completeness                      & SR                & Completeness      \\
            \midrule
            GSCE \cite{GSCE}                     & $100.0\%$                     & $100.0\%$                         & $66.7\%$          & $88.9\%$          \\
            Self-Refine \cite{SelfRefine}        & $ 100.0\%$                    & $ 100.0\%$                        & $ 50.0\%$         & $ 74.3\%$         \\
            Numerical \cite{InteractivePlanning} & $ 95.5\%$                     & $ 97.3\%$                         & $ 73.3\%$         & $ 92.4\%$         \\
            Ours                                 & $100.0\%$                     & $100.0\%$                         & $\mathbf{85.0\%}$ & $\mathbf{98.5\%}$ \\
            \bottomrule
        \end{tabular}
    \end{adjustbox}
    \label{tab:overall results}
\end{table}

For the Basic tasks, all methods showed remarkable performance. It is notable that the Numerical method \cite{InteractivePlanning} has a slightly lower performance with a $95.5\%$ SR and a $97.3\%$ completeness. This indicates that relying on numerical state observations can lead to misinterpretation in evaluation, even when handling simple tasks.

When moving to the Advanced tasks that have a higher complexity, our method still achieved a reliable performance with an $85\%$ SR and a $98.5\%$ completeness. This result shows that $98.5\%$ of required actions in all evaluated tasks are completed correctly, and $85\%$ of tasks are entirely completed without any error. As a comparison, the reliability of baseline methods drops notably when the task complexity increases due to the limitations caused by factors such as amplified bias, limited numerical reasoning capabilities of LLMs, and open-loop design without refinement. While the impact of these factors is relatively negligible when handling basic tasks, their limitations are amplified when the task complexity increases, and affect the logical reasoning capabilities of LLMs to generate operation code correctly.

\begin{table*}[!b]
    \centering
    \caption{Standard Deviation and $95\%$ Confidence Interval of Completeness for Tasks in Advanced Task Dataset}
    \begin{tabular}{l c c c c c c c c c c c c c c c }
        \toprule
        Task & \multicolumn{2}{c}{GSCE \cite{GSCE}} & \multicolumn{2}{c}{Self-Refine \cite{SelfRefine}} & \multicolumn{2}{c}{Numerical \cite{InteractivePlanning}} & \multicolumn{2}{c}{Ours} & \\
        \midrule
        & Standard & Confidence & Standard & Confidence & Standard & Confidence & Standard & Confidence \\
        & Deviation & Interval & Deviation & Interval & Deviation & Interval & Deviation & Interval \\
        \midrule
        1 & $0$ & $[1.0, 1.0]$ & $0.385$ & $[0.333, 1.0]$ & $0$ & $[1.0, 1.0]$ & $0$ & $[1.0, 1.0]$ \\
        2 & $0.128$ & $[0.778, 1.0]$ & $0.449$ & $[0.222, 1.0]$ & $0.421$ & $[0.222, 1.0]$ & $0$ & $[1.0, 1.0]$ \\
        3 & $0$ & $[1.0, 1.0]$ & $0.449$ & $[0.222, 1.0]$ & $0$ & $[1.0, 1.0]$ & $0$ & $[1.0, 1.0]$ \\
        4 & $0$ & $[1.0, 1.0]$ & $0$ & $[1.0, 1.0]$ & $0$ & $[1.0, 1.0]$ & $0$ & $[1.0, 1.0]$ \\
        5 & $0$ & $[1.0, 1.0]$ & $0$ & $[1.0, 1.0]$ & $0$ & $[1.0, 1.0]$ & $0$ & $[1.0, 1.0]$ \\
        6 & $0.173$ & $[0.700, 1.0]$ & $0.058$ & $[0.9, 1.0]$ & $0.115$ & $[0.8, 1.0]$ & $0$ & $[1.0, 1.0]$ \\
        7 & $0.167$ & $[0.444, 0.778]$ & $0.321$ & $[0.444, 1.0]$ & $0.280$ & $[0.444, 1.0]$ & $0$ & $[1.0, 1.0]$ \\
        8 & $0.096$ & $[0.778, 0.944]$ & $0.064$ & $[0.889, 1.0]$ & $0.032$ & $[0.944, 1.0]$ & $0.192$ & $[0.611, 0.944]$ \\
        9 & $0.238$ & $[0.588, 1.0]$ & $0.068$ & $[0.412, 0.529]$ & $0$ & $[1.0, 1.0]$ & $0$ & $[1.0, 1.0]$ \\
        10 & $0.513$ & $[0.111, 1.0]$ & $0$ & $[0.944, 0.944]$ & $0.289$ & $[0.5, 1.0]$ & $0$ & $[1.0, 1.0]$ \\
        11 & $0$ & $[1.0, 1.0]$ & $0$ & $[0.2, 0.2]$ & $0$ & $[1.0, 1.0]$ & $0$ & $[1.0, 1.0]$ \\
        12 & $0.052$ & $[0.909, 1.0]$ & $0.472$ & $[0.182, 1.0]$ & $0$ & $[1.0, 1.0]$ & $0$ & $[1.0, 1.0]$ \\
        13 & $0$ & $[1.0, 1.0]$ & $0.527$ & $[0.059, 1.0]$ & $0$ & $[1.0, 1.0]$ & $0$ & $[1.0, 1.0]$ \\
        14 & $0$ & $[1.0, 1.0]$ & $0.064$ & $[0.889, 1.0]$ & $0$ & $[1.0, 1.0]$ & $0$ & $[1.0, 1.0]$ \\
        15 & $0$ & $[1.0, 1.0]$ & $0.289$ & $[0.5, 1.0]$ & $0$ & $[1.0, 1.0]$ & $0$ & $[1.0, 1.0]$ \\
        16 & $0.102$ & $[0.8, 1.0]$ & $0.346$ & $[0.4, 1.0]$ & $0$ & $[1.0, 1.0]$ & $0$ & $[1.0, 1.0]$ \\
        17 & $0.058$ & $[0.9, 1.0]$ & $0$ & $[0.4, 0.4]$ & $0.173$ & $[0.6, 0.9]$ & $0.058$ & $[0.9, 1.0]$ \\
        18 & $0.229$ & $[0.579, 1.0]$ & $0$ & $[1.0, 1.0]$ & $0.169$ & $[0.684, 1.0]$ & $0.030$ & $[0.947, 1.0]$ \\
        19 & $0$ & $[1.0, 1.0]$ & $0.272$ & $[0.529, 1.0]$ & $0.238$ & $[0.588, 1.0]$ & $0$ & $[1.0, 1.0]$ \\
        20 & $0$ & $[0.944, 0.944]$ & $0$ & $[0.722, 0.722]$ & $0.032$ & $[0.944, 1.0]$ & $0$ & $[0.944, 0.944]$ \\
        \midrule
        Overall & $0.164$ & $[0.843, 0.926]$ & $0.325$ & $[0.66, 0.822]$ & $0.178$ & $[0.876, 0.964]$ & $\mathbf{0.055}$ & $\mathbf{[0.969, 0.995]}$ \\
        \bottomrule
    \end{tabular}
    \label{tab:advanced task deviation and CI}
\end{table*}

\underline{Statistical Analysis of Completeness}:
To evaluate the stability and consistency of our method, we performed statistical analysis for the completeness of advanced tasks. Specifically, the standard deviation and the $95\%$ confidence interval are measured for each advanced task across all three repetitions. As summarized in Table~\ref{tab:advanced task deviation and CI}, our method has a low overall standard deviation of $0.055$ and a high overall confidence interval $[0.969, 0.995]$. These values also remain stable among almost all advanced tasks except for one of the failure cases. A more detailed analysis of the failure cases is also provided below. Compared with the baseline methods, which have higher standard deviations and wider confidence intervals for most tasks as shown in Table~\ref{tab:advanced task deviation and CI}, our method clearly achieves better stability and consistency.

\begin{figure*}[!thbp]
    \centering{\includegraphics[scale=0.59]{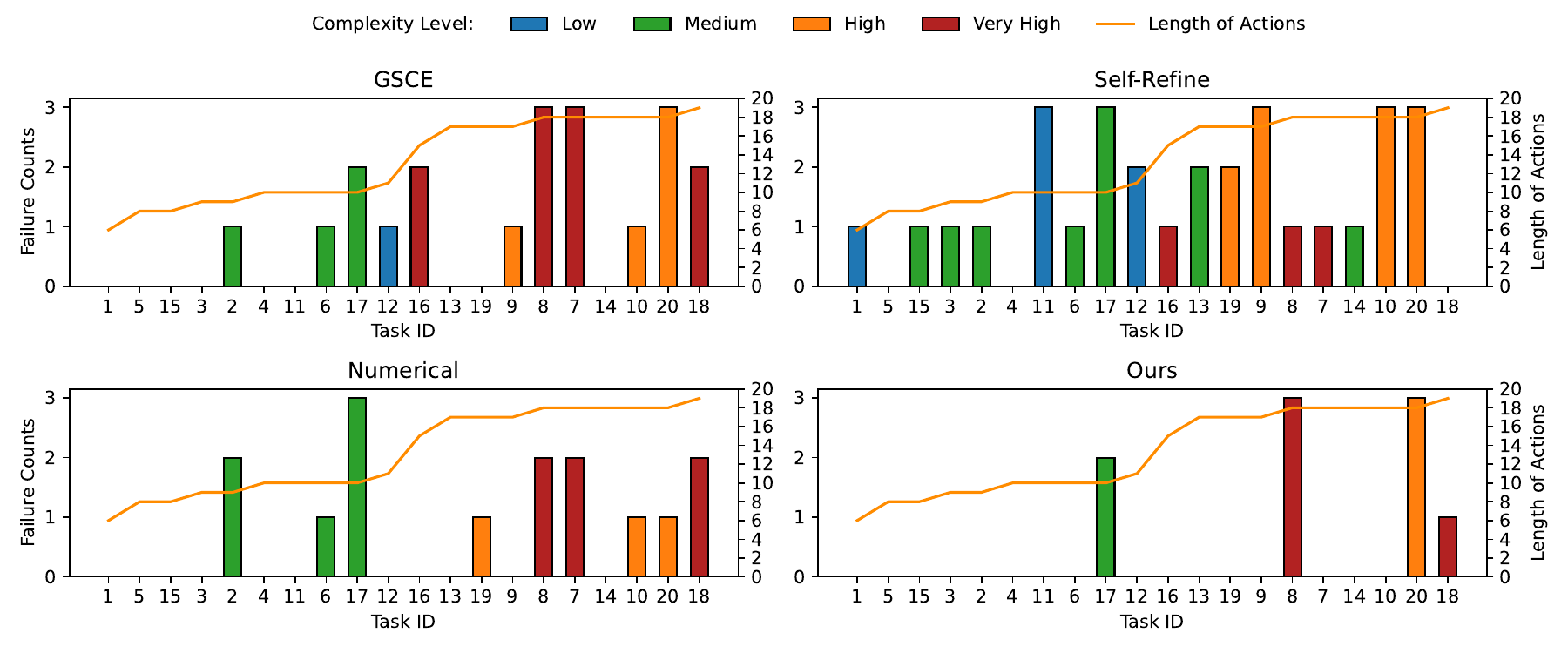}}
    \caption{Failure counts of different methods for each advanced task at different complexity levels.} 
    \label{fig:failed count}
\end{figure*}

\begin{figure*}[!thbp]
    \centering{\includegraphics[scale=0.59]{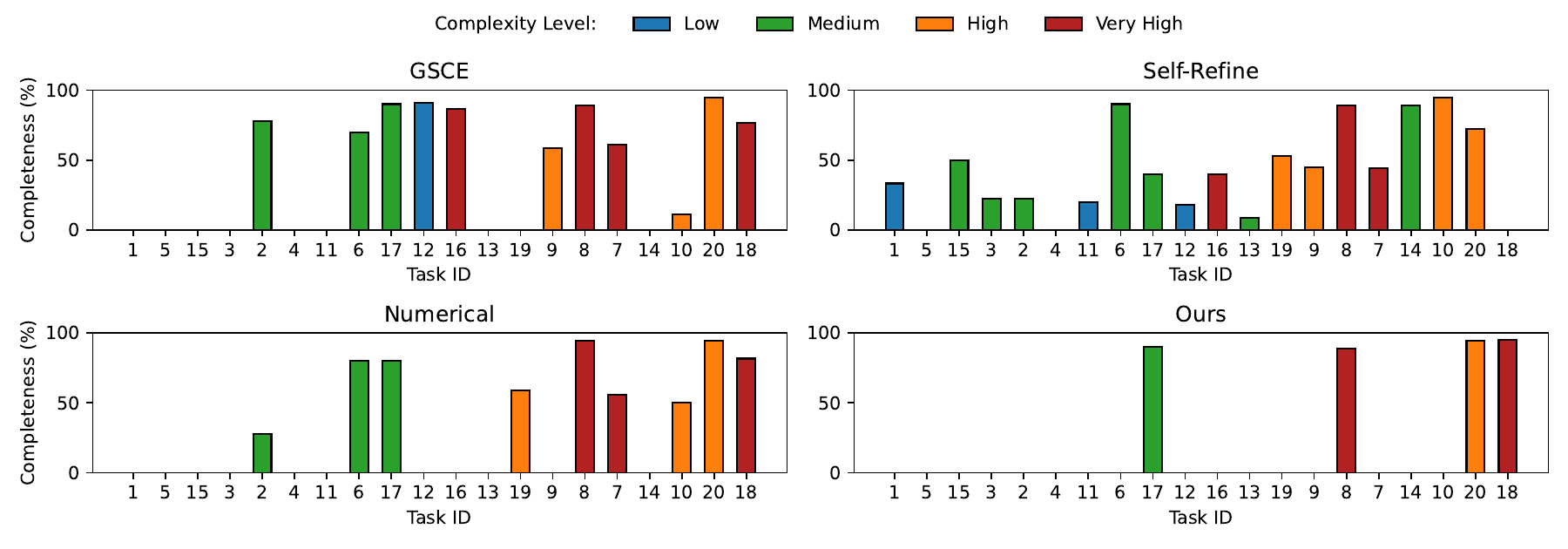}}
    \caption{Completeness of each failed task of different methods at different complexity levels}
    \label{fig:failed completeness}
\end{figure*}

\begin{table}[!thbp] 
    \centering
    \caption{Failure Cases Completeness in Each Level of Complexity}
    \begin{tabular}{l c c c c c}
        \toprule
        & Low & Medium & High & Very   & Overall \\
        &   &   &   &   High &   \\
        \midrule
        GSCE \cite{GSCE} & $90.9\%$ & $81.9\%$ & $70.7\%$ & $77.6\%$ & $ 77.4\%$\\
        Self-Refine \cite{SelfRefine} & $21.6\%$ & $41.1\%$ & $67.4\%$ & $57.8\%$ & $ 48.5\%$ \\
        Numerical \cite{InteractivePlanning} & --- & $62.6\%$ & $67.8\%$ & $77.2\%$ & $ 69.5\%$ \\
        Ours & --- & $\mathbf{90.0\%}$ & $\mathbf{94.4\%}$ & $\mathbf{90.4\%}$ & $\mathbf{91.6\%}$ \\
        \bottomrule
    \end{tabular}
    \label{tab:failed complexity}
\end{table}

\underline{Failure Case Analysis}: To uncover the detailed performance of each method in its failed advanced tasks, we performed an in-depth failure case analysis. In particular, we aim to understand how many actions are correctly completed when a task fails. As summarized in Table \ref{tab:failed complexity}, our method achieves a high average completeness of $91.6\%$ for all failed tasks, i.e., the operation code generated by our method can still complete $91.6\%$ of the required actions in the task and remains highly aligned with the task objectives even when errors occur. Differently, the average completeness of the compared baseline methods ranges from $48.5\%$ to $77.4\%$ for their failed tasks, which indicates errors occur for a significant portion of the required actions in these tasks. Fig.~\ref{fig:failed count} further details the failure count on each failed task of all methods in our three rounds of experiment with task complexity and length of actions information. Our method has failures on only four tasks during three rounds of experiments. As a comparison, the baseline methods failed 11 (GSCE \cite{GSCE}), 17 (Self-Refine \cite{SelfRefine}), and 9 (Numerical \cite{InteractivePlanning}) tasks, respectively. In addition, Table \ref{tab:failed complexity} and Fig.~\ref{fig:failed completeness} also show that our method has a stable high completeness of $90\%+$ for all failed tasks at different complexity levels. In contrast, the baseline methods have not only significantly lower average completeness ($21\%$ to $90.9\%$) on failed tasks at different levels but also a large variation in performance of different tasks.

Based on our evaluation results, our method achieves a higher reliability compared with the baseline methods, not only because of its high confidence in generating fully correct operation code, but also its high completeness of the required actions of tasks when a small amount of errors occur.

\begin{table*}[!htbp]
    \centering
    \caption{Evaluator Configurations Description and Example System Prompts }
    \begin{tabular}{L{3.8cm} p{13cm}}
        \toprule
        \textbf{Evaluator Configuration} & \multicolumn{1}{c}{\textbf{Description}}\\
        \specialrule{.5pt}{2.5pt}{.4pt} 
        (1) Roles & This configuration defines the role for the LLM as the code evaluator, and its goal is to evaluate whether the UAV flight trajectory is aligned with the task description and output standardized feedback for the code generator.\\
        \midrule
        (2) Roles + Rules & Besides the roles in (1), this configuration also includes specific rules for the LLM to follow. The rules introduce explicit evaluation criteria on the UAV's actions, spatial positions, flight paths, coordinate system, and offset tolerance that direct the LLM's evaluation of the UAV trajectory. \\
        \midrule
        (3) Roles + References & Besides the roles in (1), this configuration also provides reference examples to support evaluation and generate feedback.  These references include pairs of task descriptions and corresponding semantic trajectory observations.\\
        \midrule
        (4) Roles + Rules + References & This configuration combines roles, rules, and references to provide comprehensive support information for the LLM to perform its job correctly.\\
        \specialrule{.8pt}{.5pt}{2.5pt} 
        \textbf{} & \multicolumn{1}{c}{\textbf{Example System Prompts}} \\
        \midrule
        Roles & 1. You will reason about the difference between the task description and drone actions. \\
        & 2. If the task description and drone actions are aligned, you should output ``YES'' only. Otherwise, you should output ``NO''; please output where the drone actions mismatch the task description if your answer is ``NO''. \\
        \midrule 
        Rules  & 1. For tasks that require perpendicularly examining a square area, you must check whether the drone is oriented perpendicularly inwards the square interior; facing toward the exterior is not allowed. \\
         & 2. You should reason about the spatial locations of square patterns from actions and ensure the drone flies the square pattern in the correct spatial position as outlined in the task description.\\
         & 3. The drone uses the NED coordinate system (world coordinates) represented as [x, y, z], the positive X axis is North/forward, the positive Y axis is East/right, and the negative Z axis is up/climb.\\
        \midrule
        References & Here is a correct path of a 4-meter square pattern (forward/right/backward/left) for reference: 1. Fly 4 meters North in the world axis while facing north. The drone moves to [4, 0]. 2. Fly 4 meters East in the world axis while facing north. The drone moves to [4, 4]. 3. Fly 4 meters South in the world axis while facing north. The drone moves to [0, 4]. 4. Fly 4 meters West in the world axis while facing north. The drone moves to [0, 0]. \\
        
        \midrule
        \multicolumn{2}{l}{*Note: more examples of system prompts for roles, rules, and references are available at the project website: \url{https://github.com/ai-uavsec/CLGSCE.}} \\
    \end{tabular}
    \label{tab:evaluator setup}
\end{table*}

\subsubsection{\textbf{Number of Refinement Iterations}} \label{sec:num of itr}
In our proposed framework, the operation code generated by the Code Generator LLM will be refined through multiple iterations. The increased rounds of iterations provide additional opportunities to improve the code quality. However, they also introduce additional system cost (e.g., LLM API cost and processing time). To balance this tradeoff and identify an appropriate number of iterations, we performed experiments of our framework on advanced tasks with $0$ to $10$ rounds of iterations. As described in Algorithm~\ref{alg:overall}, the early stopping is used in our design when the evaluator outputs ``YES'', which prevents additional unnecessary refinement iterations.

\begin{figure}[!htbp]
    \centering
    \includegraphics[width=0.48\textwidth]{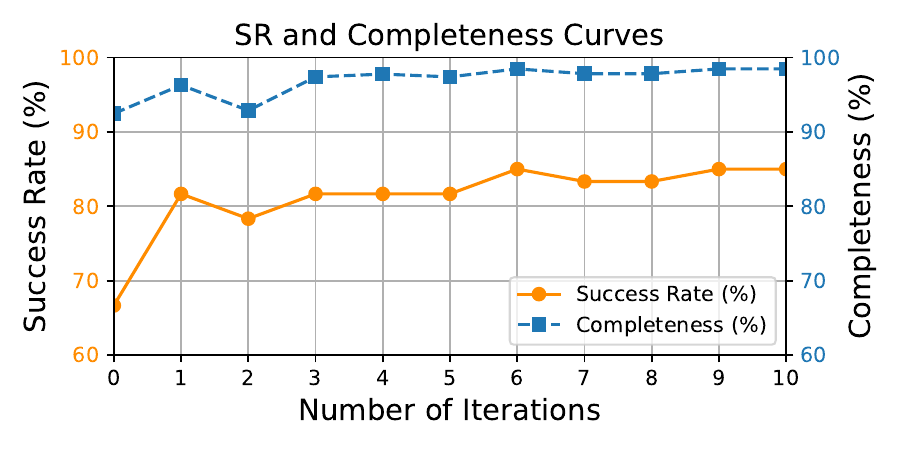}
    \caption{Curves of SR (left) and completeness (right) over the number of refinement iterations for our method on advanced tasks.}
    \label{fig:number of itr}
\end{figure}

\begin{figure}[!htbp]
    \centering
    \includegraphics[width=0.48\textwidth]{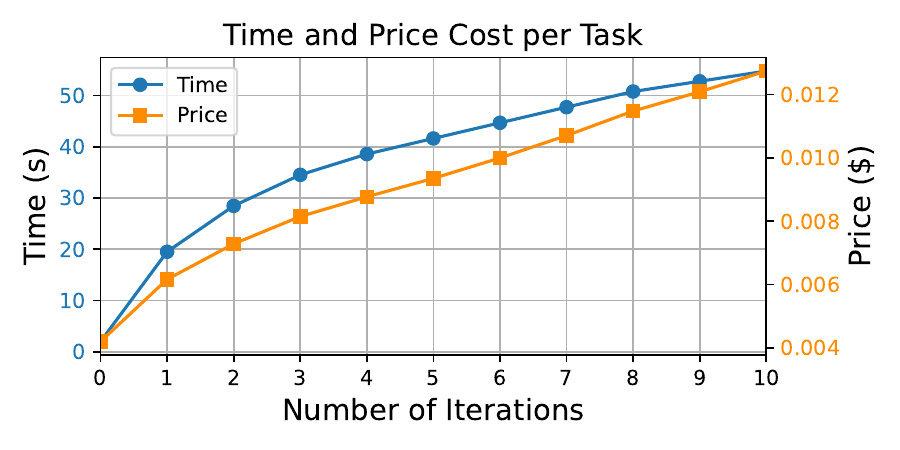}
    \caption{Curves of per-task time and price over the number of refinement iterations.}
    \label{fig:time price}
\end{figure}

The mean of our three repetition results is presented in Fig.~\ref{fig:number of itr}, which shows the changes of SR and completeness with the increment of refinement iterations. Specifically, the mean iteration for our method to start converging with a relatively stable performance is 3. With a mean iteration of 6, our method is converged with the best performance in terms of both completeness and success rate. The performance declines slightly after the $6\textsuperscript{th}$ iteration because LLMs can exhibit sycophantic behavior and tend to adapt or change outputs to appear agreeable \cite{flipflop, LLM-behaviors} when receiving a negative query, and thus may perform unnecessary changes after sufficient refinement has been performed. This observation demonstrates that iterative refinement can effectively contribute to the LLM-driven UAV operation code generation. Meanwhile, it is also necessary to impose an upper bound on refinement iterations to balance the tradeoff between the success rate and system cost. As presented in Fig.~\ref{fig:time price}, the computational cost and OpenAI API usage cost increase with the number of iterations. Thus, it is unnecessary to continue the refinement when the gain of success rate becomes very limited, i.e., after $6$ iterations in our setting. The number of iterations can be adjusted for different types of tasks. For example, setting a lower number when the tasks to complete have a relatively low complexity.

\subsection{Evaluator Configuration Analysis} \label{sec:evaluator config}
As the accuracy of the evaluation feedback directly influences the quality of code refinement, we analyze how the evaluator's design influences both evaluation accuracy and overall system performance. We evaluated four configurations of the evaluator's system prompt, including (1) roles, (2) roles $+$ rules, (3) roles $+$ references, and (4) roles $+$ rules $+$ references. As specified in Section~\ref{sec:evaluation}, the \emph{roles} define the LLM to be an evaluator and serve as the essential component to ensure the LLM will perform the evaluation task. Thus, we include the \emph{roles} in each configuration of the evaluator. For each configuration, we instantiate the evaluator's system prompt by including only the indicated components. The detailed description of each evaluator configuration with sample system prompts for roles, rules, and references is provided in Table \ref{tab:evaluator setup}.

\begin{table}[!htbp]
    \centering
    \caption{Performance over Evaluator Configurations}
    \begin{tabular}{lcccc}
        \toprule
                 & (1)        & (2)        & (3)        & (4)             \\
        \midrule
        Roles                 & \checkmark & \checkmark & \checkmark & \checkmark        \\
        Rules                 &            & \checkmark &            & \checkmark        \\
        References            &            &            & \checkmark & \checkmark        \\
        \midrule
        SR & $75.0\%$ & $81.7\%$ & $81.7\%$ & $\mathbf{85.0\%}$\\
        Completeness & $92.7\%$ & $92.8\%$ & $93.8\%$ & $\mathbf{98.5\%}$\\
        \bottomrule
    \end{tabular}
    \label{tab:evaluator configuration}
\end{table}

Table~\ref{tab:evaluator configuration} presents the overall performance of each configuration in terms of SR and completeness on advanced tasks. Given the fact that the \emph{roles} provide the essential information that helps the LLM understand its role as the evaluator, the roles-only evaluator configuration can already achieve a $75\%$ SR and $92.7\%$ completeness. When adding the \emph{rules} or the \emph{references} components to the evaluator configuration, they all improve the overall performance of our framework with an increased SR of $81.7\%$. These noticeable improvements suggest that explicit evaluation criteria (rules) and example demonstrations (references) each enhance evaluation quality. When all three components are used in the evaluator configuration, the overall performance of our framework is further improved and reaches $85.0\%$ SR and $98.5\%$ completeness, proving this configuration enables the evaluator to provide accurate feedback that guides more reliable code refinement. Therefore, the mission-related context contextual information from roles, rules, and references can effectively contribute to the evaluator's capability toward UAV operation code analysis.

\begin{figure}[!htbp]
    \centering
    \includegraphics[width=0.48\textwidth]{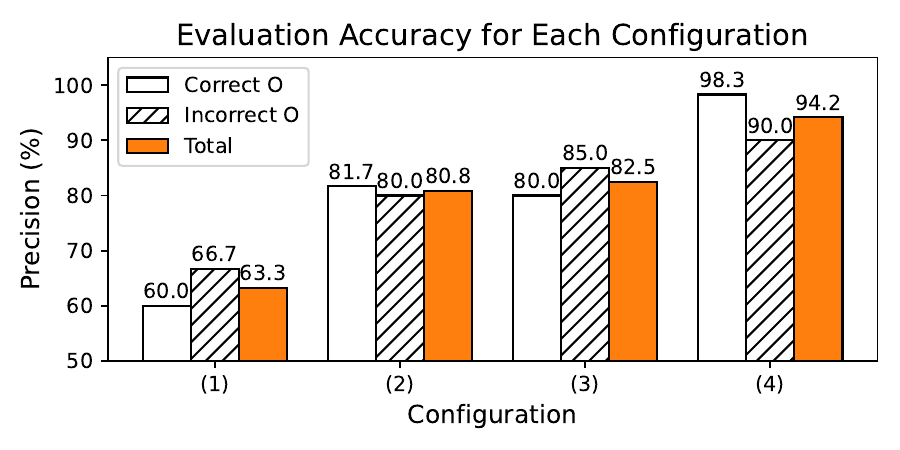}
    \caption{Evaluation accuracy of different evaluator configurations.}
    \label{fig:evaluator configuration}
\end{figure}

\underline{Evaluation Accuracy}: To further understand the impact of different evaluator configurations, we also analyze their evaluation performance in terms of accuracy. For each advanced task in our experiment, we build a corresponding correct semantic trajectory observation (Correct O) and an incorrect semantic trajectory observation (Incorrect O). The evaluation accuracy of each evaluator configuration is measured by the percentage of correctly evaluated semantic trajectory observations, i.e., output correct for Correct O and incorrect for Incorrect O. 
The results in Fig.~\ref{fig:evaluator configuration} show that the evaluation accuracy improves with the inclusion of more contextual information in the evaluator's configuration. Specifically, using roles only yields the lowest accuracy, indicating that merely defining the LLM's role as an evaluator is inadequate for reliably assessing trajectory correctness or identifying deviations. The inclusion of rules leads to a substantial improvement, suggesting the importance of providing explicit evaluation criteria to the LLM. Similarly, incorporating references enhances the evaluator's ability to identify incorrect outputs via LLM's in-context learning~\cite{URIAL}. However, these configurations remain below $90\%$ accuracy. In contrast, the configuration that combines roles, rules, and references yields the highest performance, which achieves $98.3\%$ accuracy on Correct O, $90\%$ accuracy on Incorrect O, and $94.2\%$ for the total accuracy. These results highlight the effectiveness of combining explicit evaluation criteria with reference examples to maximize the accuracy of LLM evaluation. Furthermore, the near-perfect accuracy for Correct O can minimize the unnecessary refinement for the correct operation code.

\section{Discussion and Future Work}\label{sec:discussion}
\subsection{Adaptability to Other Domains}
While the proposed LLM-driven operation code generation framework in this paper is originally designed for UAVs, it also has the potential to be adapted to other robotic platforms, such as ground robots, underwater vehicles, and robot arms. To apply our framework to other domains, two key adjustments in terms of system prompts and simulation-based evaluation will need to be performed based on the domain-specific knowledge. Specifically, while some general system prompts from our framework that define the roles of the LLMs as the code generator and evaluator can be used for other domains, domain-specific knowledge will be needed to update the rules and references. For example, the rule currently used in our design to regulate the usage of the NED coordinate system may not be applicable to non-aerospace applications. In addition, our framework enables a simulation-based operation code execution and state observation by tailoring the UAV simulation tool AirSim. When applying to other domains, a corresponding simulation platform will also be needed and customized to support the data format of our proposed state transformation algorithm. Currently, most major robotic platforms have well-established open-source simulation tools \cite{robot-simulation-platforms}, and hence make the potential adaptation of our framework feasible with domain-specific adjustment.

\subsection{Limitations and Challenges}
While our proposed framework has demonstrated promising results, it still faces challenges and limitations. First, it currently depends on the cloud-based LLM models, which may introduce latency for real-time UAV operations. Moreover, querying cloud-based LLM APIs requires a reliable internet connection, which may not always be available in complex or remote operating environments. Furthermore, the overall performance of the framework is built upon the capability of the large-scale foundational LLMs (e.g., OpenAI models). Adapting the framework to smaller language models may require additional prompt engineering or fine-tuning to achieve reliable performance.

\subsection{Future Work}
In our future research, we will further explore the generality of our proposed framework for more robotic platforms. In addition, aligning with existing research \cite{incoro, InteractivePlanning, CoPAL, CAPE, SelfRefine, trust}, our current design focuses on generating the complete operation code for the given task and then deploying it on the UAV for execution. Considering the potential changes of the execution plan due to different factors, our future research will also explore how to support real-time interactive UAV operation code generation powered by LLMs. Moreover, we will also explore the possibility of enabling on-board deployment of energy-efficient LLM models (e.g., RWKV \cite{RWKV-7}) to enable more intelligent real-time UAV controls. Given the restricted battery and computing resources of UAVs, the key challenge that needs to be addressed is how to balance the resource consumption of LLMs and their performance in terms of efficiency and reliability.

\section{Conclusion}\label{sec:conclusion}
In this paper, we introduce an LLM-driven closed-loop feedback and refinement framework that leverages semantic observations for UAV operation code generation. Our framework comprises two LLM components: a Code Generator, which generates control code; and an Evaluator, which analyzes semantic trajectory observations rather than numerical states. By iteratively feeding the evaluation feedback back into the Code Generator, our method corrects errors in simulation before real-world deployment. Experimental results demonstrate that semantic trajectory feedback markedly enhances the Evaluator's ability to detect deviations and deliver precise corrective feedback. As a result, our framework achieves reliable LLM-driven UAV operation code generation and significantly outperforms baseline methods when the task complexity increases. 

\bibliographystyle{IEEEtran}
\bibliography{ref.bib}

@article{few-shot,
  title={Language models are few-shot learners},
  author={Brown, Tom and Mann, Benjamin and Ryder, Nick and Subbiah, Melanie and Kaplan, Jared D and Dhariwal, Prafulla and Neelakantan, Arvind and Shyam, Pranav and Sastry, Girish and Askell, Amanda and others},
  journal={Advances in neural information processing systems},
  volume={33},
  pages={1877--1901},
  year={2020}
}

@inproceedings{LLM-behaviors,
  title={Discovering language model behaviors with model-written evaluations},
  author={Perez, Ethan and Ringer, Sam and Lukosiute, Kamile and Nguyen, Karina and Chen, Edwin and Heiner, Scott and Pettit, Craig and Olsson, Catherine and Kundu, Sandipan and Kadavath, Saurav and others},
  booktitle={Findings of the association for computational linguistics: ACL 2023},
  pages={13387--13434},
  year={2023}
}

@article{flipflop,
  title={Are you sure? challenging llms leads to performance drops in the flipflop experiment},
  author={Laban, Philippe and Murakhovs' ka, Lidiya and Xiong, Caiming and Wu, Chien-Sheng},
  journal={arXiv preprint arXiv:2311.08596},
  year={2023}
}

@article{preemptive,
  title={Preemptive detection and correction of misaligned actions in llm agents},
  author={Fang, Haishuo and Zhu, Xiaodan and Gurevych, Iryna},
  journal={arXiv preprint arXiv:2407.11843},
  year={2024}
}

@ARTICLE{uav_iot_applications,
  author={Cheng, Nan and Wu, Shen and Wang, Xiucheng and Yin, Zhisheng and Li, Changle and Chen, Wen and Chen, Fangjiong},
  journal={IEEE Internet of Things Journal}, 
  title={AI for UAV-Assisted IoT Applications: A Comprehensive Review}, 
  year={2023},
  volume={10},
  number={16},
  pages={14438-14461},
  doi={10.1109/JIOT.2023.3268316}}

@article{ChatGPTRobotics,
  author   = {Vemprala, Sai H. and Bonatti, Rogerio and Bucker, Arthur and Kapoor, Ashish},
  journal  = {IEEE Access},
  title    = {ChatGPT for Robotics: Design Principles and Model Abilities},
  year     = {2024},
  volume   = {12},
  number   = {},
  pages    = {55682-55696},
  doi      = {10.1109/ACCESS.2024.3387941}
}

@article{TypeFly,
  title   = {Typefly: Flying drones with large language model},
  author  = {Chen, Guojun and Yu, Xiaojing and Ling, Neiwen and Zhong, Lin},
  journal = {arXiv preprint arXiv:2312.14950},
  year    = {2023}
}

@inproceedings{CodeasPolicies,
  title        = {Code as policies: Language model programs for embodied control},
  author       = {Liang, Jacky and Huang, Wenlong and Xia, Fei and Xu, Peng and Hausman, Karol and Ichter, Brian and Florence, Pete and Zeng, Andy},
  booktitle    = {2023 IEEE International Conference on Robotics and Automation (ICRA)},
  pages        = {9493--9500},
  year         = {2023},
  organization = {IEEE}
}

@article{URIAL,
  title   = {The unlocking spell on base llms: Rethinking alignment via in-context learning},
  author  = {Lin, Bill Yuchen and Ravichander, Abhilasha and Lu, Ximing and Dziri, Nouha and Sclar, Melanie and Chandu, Khyathi and Bhagavatula, Chandra and Choi, Yejin},
  journal = {arXiv preprint arXiv:2312.01552},
  year    = {2023}
}

@inproceedings{PromptBook,
  author    = {Arenas, Montserrat Gonzalez and Xiao, Ted and Singh, Sumeet and Jain, Vidhi and Ren, Allen and Vuong, Quan and Varley, Jake and Herzog, Alexander and Leal, Isabel and Kirmani, Sean and Prats, Mario and Sadigh, Dorsa and Sindhwani, Vikas and Rao, Kanishka and Liang, Jacky and Zeng, Andy},
  booktitle = {2024 IEEE International Conference on Robotics and Automation (ICRA)},
  title     = {How to Prompt Your Robot: A PromptBook for Manipulation Skills with Code as Policies},
  year      = {2024},
  volume    = {},
  number    = {},
  pages     = {4340-4348},
  doi       = {10.1109/ICRA57147.2024.10610784}
}

@inproceedings{ISR-LLM,
  author    = {Zhou, Zhehua and Song, Jiayang and Yao, Kunpeng and Shu, Zhan and Ma, Lei},
  booktitle = {2024 IEEE International Conference on Robotics and Automation (ICRA)},
  title     = {ISR-LLM: Iterative Self-Refined Large Language Model for Long-Horizon Sequential Task Planning},
  year      = {2024},
  volume    = {},
  number    = {},
  pages     = {2081-2088},
  doi       = {10.1109/ICRA57147.2024.10610065}
}

@article{CoT,
  title   = {Chain-of-thought prompting elicits reasoning in large language models},
  author  = {Wei, Jason and Wang, Xuezhi and Schuurmans, Dale and Bosma, Maarten and Xia, Fei and Chi, Ed and Le, Quoc V and Zhou, Denny and others},
  journal = {Advances in neural information processing systems},
  volume  = {35},
  pages   = {24824--24837},
  year    = {2022}
}

@inproceedings{AirSim,
  title        = {Airsim: High-fidelity visual and physical simulation for autonomous vehicles},
  author       = {Shah, Shital and Dey, Debadeepta and Lovett, Chris and Kapoor, Ashish},
  booktitle    = {Field and Service Robotics: Results of the 11th International Conference},
  pages        = {621--635},
  year         = {2018},
  organization = {Springer}
}

@inproceedings{PromptEngineering,
  title        = {Prompt engineering in large language models},
  author       = {Marvin, Ggaliwango and Hellen, Nakayiza and Jjingo, Daudi and Nakatumba-Nabende, Joyce},
  booktitle    = {International conference on data intelligence and cognitive informatics},
  pages        = {387--402},
  year         = {2023},
  organization = {Springer}
}

@article{UAVsmeetllmsoverviews,
author = {Tian, Yonglin and Lin, Fei and Li, Yiduo and Zhang, Tengchao and Zhang, Qiyao and Fu, Xuan and Huang, Jun and Dai, Xingyuan and Wang, Yutong and Tian, Chunwei and Li, Bai and Lv, Yisheng and Kov\'{a}cs, Levente and Wang, Fei-Yue},
title = {UAVs meet LLMs: Overviews and perspectives towards agentic low-altitude mobility},
year = {2025},
issue_date = {Oct 2025},
publisher = {Elsevier Science Publishers B. V.},
address = {NLD},
volume = {122},
number = {C},
issn = {1566-2535},
doi = {10.1016/j.inffus.2025.103158},
journal = {Inf. Fusion},
month = jun,
numpages = {33}
}

@inproceedings{CoPAL,
  author    = {Joublin, Frank and Ceravola, Antonello and Smirnov, Pavel and Ocker, Felix and Deigmoeller, Joerg and Belardinelli, Anna and Wang, Chao and Hasler, Stephan and Tanneberg, Daniel and Gienger, Michael},
  booktitle = {2024 IEEE International Conference on Robotics and Automation (ICRA)},
  title     = {CoPAL: Corrective Planning of Robot Actions with Large Language Models},
  year      = {2024},
  volume    = {},
  number    = {},
  pages     = {8664-8670},
  doi       = {10.1109/ICRA57147.2024.10610434}
}

@inproceedings{CLEAR,
  title     = {Language, Camera, Autonomy! Prompt-engineered Robot Control for Rapidly Evolving Deployment},
  author    = {Macdonald, Jacob P and Mallick, Rohit and Wollaber, Allan B and Pe{\~n}a, Jaime D and McNeese, Nathan and Siu, Ho Chit},
  booktitle = {Companion of the 2024 ACM/IEEE International Conference on Human-Robot Interaction},
  pages     = {717--721},
  year      = {2024}
}

@inproceedings{GSCE,
  author    = {Wang, Wenhao and Li, Yanyan and Jiao, Long and Yuan, Jiawei},
  booktitle = {2025 International Conference on Unmanned Aircraft Systems (ICUAS)},
  title     = {GSCE: a Prompt Framework With Enhanced Reasoning for Reliable LLM-Driven Drone Control},
  year      = {2025},
  volume    = {},
  number    = {},
  pages     = {441-448},
  doi       = {10.1109/ICUAS65942.2025.11007864}
}

@inproceedings{InteractivePlanning,
  author    = {Sun, Lingfeng and Jha, Devesh K. and Hori, Chiori and Jain, Siddarth and Corcodel, Radu and Zhu, Xinghao and Tomizuka, Masayoshi and Romeres, Diego},
  booktitle = {2024 IEEE International Conference on Robotics and Automation (ICRA)},
  title     = {Interactive Planning Using Large Language Models for Partially Observable Robotic Tasks},
  year      = {2024},
  volume    = {},
  number    = {},
  pages     = {14054-14061},
  doi       = {10.1109/ICRA57147.2024.10610981}
}

@inproceedings{constraints,
  title     = {How You Prompt Matters! {E}ven Task-Oriented Constraints in Instructions Affect {LLM}-Generated Text Detection},
  author    = {Koike, Ryuto  and
               Kaneko, Masahiro  and
               Okazaki, Naoaki},
  editor    = {Al-Onaizan, Yaser  and
               Bansal, Mohit  and
               Chen, Yun-Nung},
  booktitle = {Findings of the Association for Computational Linguistics: EMNLP 2024},
  month     = nov,
  year      = {2024},
  address   = {Miami, Florida, USA},
  publisher = {Association for Computational Linguistics},
  url       = {https://aclanthology.org/2024.findings-emnlp.841/},
  doi       = {10.18653/v1/2024.findings-emnlp.841},
  pages     = {14384--14395}
}

@electronic{o3mini,
  author = {{OpenAI Models}},
  title  = {o3-mini},
  year   = {2025},
  note   = {Accessed: 25-May-2025},
  url    = {https://platform.openai.com/docs/models/o3-mini}
}

@article{deepseek,
  title   = {Deepseek-v3 technical report},
  author  = {Liu, Aixin and Feng, Bei and Xue, Bing and Wang, Bingxuan and Wu, Bochao and Lu, Chengda and Zhao, Chenggang and Deng, Chengqi and Zhang, Chenyu and Ruan, Chong and others},
  journal = {arXiv preprint arXiv:2412.19437},
  year    = {2024}
}

@inproceedings{trust,
  title     = {Trust the {PR}oC3S: Solving Long-Horizon Robotics Problems with {LLM}s and Constraint Satisfaction},
  author    = {Aidan Curtis and Nishanth Kumar and Jing Cao and Tom{\'a}s Lozano-P{\'e}rez and Leslie Pack Kaelbling},
  booktitle = {8th Annual Conference on Robot Learning},
  year      = {2024},
}

@inproceedings{CAPE,
  author    = {Raman, Shreyas Sundara and Cohen, Vanya and Idrees, Ifrah and Rosen, Eric and Mooney, Raymond and Tellex, Stefanie and Paulius, David},
  booktitle = {2024 IEEE International Conference on Robotics and Automation (ICRA)},
  title     = {CAPE: Corrective Actions from Precondition Errors using Large Language Models},
  year      = {2024},
  volume    = {},
  number    = {},
  pages     = {14070-14077},
  doi       = {10.1109/ICRA57147.2024.10611376}
}

@inproceedings{OpenLoopQuestions,
  author    = {Shao, Hao and Hu, Yuxuan and Wang, Letian and Song, Guanglu and Waslander, Steven L. and Liu, Yu and Li, Hongsheng},
  booktitle = {2024 IEEE/CVF Conference on Computer Vision and Pattern Recognition (CVPR)},
  title     = {LMDrive: Closed-Loop End-to-End Driving with Large Language Models},
  year      = {2024},
  volume    = {},
  number    = {},
  pages     = {15120-15130},
  doi       = {10.1109/CVPR52733.2024.01432}
}

@article{SelfRefine,
  title   = {Self-refine: Iterative refinement with self-feedback},
  author  = {Madaan, Aman and Tandon, Niket and Gupta, Prakhar and Hallinan, Skyler and Gao, Luyu and Wiegreffe, Sarah and Alon, Uri and Dziri, Nouha and Prabhumoye, Shrimai and Yang, Yiming and others},
  journal = {Advances in Neural Information Processing Systems},
  volume  = {36},
  pages   = {46534--46594},
  year    = {2023}
}

@inproceedings{LLM-bias,
  title     = {Pride and Prejudice: {LLM} Amplifies Self-Bias in Self-Refinement},
  author    = {Xu, Wenda  and
               Zhu, Guanglei  and
               Zhao, Xuandong  and
               Pan, Liangming  and
               Li, Lei  and
               Wang, William},
  booktitle = {Proceedings of the 62nd Annual Meeting of the Association for Computational Linguistics (Volume 1: Long Papers)},
  year      = {2024},
  doi       = {10.18653/v1/2024.acl-long.826},
  pages     = {15474--15492}
}

@article{NumberUnderstanding,
  title   = {Number Cookbook: Number Understanding of Language Models and How to Improve It},
  author  = {Yang, Haotong and Hu, Yi and Kang, Shijia and Lin, Zhouchen and Zhang, Muhan},
  journal = {arXiv preprint arXiv:2411.03766},
  year    = {2024}
}

@article{incoro,
  title   = {Incoro: In-context learning for robotics control with feedback loops},
  author  = {Zhu, Jiaqiang Ye and Cano, Carla Gomez and Bermudez, David Vazquez and Drozdzal, Michal},
  journal = {arXiv preprint arXiv:2402.05188},
  year    = {2024}
}

@article{LLMrandomness,
  title={LLM Stability: A detailed analysis with some surprises},
  author={Atil, Berk and Chittams, Alexa and Fu, Liseng and Ture, Ferhan and Xu, Lixinyu and Baldwin, Breck},
  journal={arXiv preprint arXiv:2408.04667},
  year={2024}
}

@misc{uav-crash,
      title={When Uncertainty Leads to Unsafety: Empirical Insights into the Role of Uncertainty in Unmanned Aerial Vehicle Safety}, 
      author={Sajad Khatiri and Fatemeh Mohammadi Amin and Sebastiano Panichella and Paolo Tonella},
      year={2025},
      eprint={2501.08908},
      archivePrefix={arXiv},
      primaryClass={cs.SE},
      url={https://arxiv.org/abs/2501.08908}, 
}

@inproceedings{robot-simulation-platforms,
author = {Staranowicz, Aaron and Mariottini, Gian Luca},
title = {A survey and comparison of commercial and open-source robotic simulator software},
year = {2011},
isbn = {9781450307727},
publisher = {Association for Computing Machinery},
address = {New York, NY, USA},
booktitle = {Proceedings of the 4th International Conference on PErvasive Technologies Related to Assistive Environments},
articleno = {56},
numpages = {8},
location = {Heraklion, Crete, Greece},
series = {PETRA '11}
}

@misc{RWKV-7,
      title={RWKV-7 "Goose" with Expressive Dynamic State Evolution}, 
      author={Bo Peng and Ruichong Zhang and Daniel Goldstein and Eric Alcaide and Xingjian Du and Haowen Hou and Jiaju Lin and Jiaxing Liu and Janna Lu and William Merrill and Guangyu Song and Kaifeng Tan and Saiteja Utpala and Nathan Wilce and Johan S. Wind and Tianyi Wu and Daniel Wuttke and Christian Zhou-Zheng},
      year={2025},
      eprint={2503.14456},
      archivePrefix={arXiv},
      primaryClass={cs.CL},
      url={https://arxiv.org/abs/2503.14456}, 
}

\end{document}